\documentclass[11pt]{article}
\usepackage{apacite}
\usepackage[utf8]{inputenc}
\usepackage[a4paper, total={6in, 8in}]{geometry}
\usepackage{parskip}
\usepackage{amsmath}
\usepackage{amsfonts}
\usepackage{amssymb}
\usepackage{url} 
\usepackage{graphicx}
\usepackage{svg}
\usepackage{placeins}
\usepackage{enumerate}
\usepackage{array}
\usepackage{color}
\usepackage{balance}

\newcommand{\encoding}{E^*} % \mathcal{E}
\newcommand{\prototype}{q}

\title{Word reuse and combination support efficient communication of emerging concepts}
\author{
  Aotao Xu$^{1,2}$, Charles Kemp$^3$, Lea Frermann$^2$, Yang Xu$^{1,4}$ \\
  {\small $^1$Department of Computer Science, University of Toronto} \\
  {\small $^2$School of Computing and Information Systems, University of Melbourne} \\
  {\small $^3$School of Psychological Sciences, University of Melbourne} \\
  {\small $^4$Cognitive Science Program, University of Toronto}
}
\date{}

\begin{document}

\maketitle

\begin{abstract}
A key function of the lexicon is to express novel concepts as they emerge over time through a process known as lexicalization. The most common lexicalization strategies are the reuse and combination of existing words, but they have typically been studied separately in the areas of word meaning extension and word formation. Here we offer an information-theoretic account of how both strategies are constrained by a fundamental tradeoff between competing communicative pressures: word reuse tends to preserve the average length of word forms at the cost of less precision, while word combination tends to produce more informative words at the expense of greater word length. We test our proposal against a large dataset of reuse items and compounds that appeared in English, French and Finnish over the past century. We find that these historically emerging items achieve higher levels of communicative efficiency than hypothetical ways of constructing the lexicon, and both literal reuse items and compounds tend to be more efficient than their non-literal counterparts. These results suggest that reuse and combination are both consistent with a unified account of lexicalization grounded in the theory of efficient communication.

\end{abstract}

\section{Introduction}
The human lexicon is not static but evolves over time. A central role of the lexicon in this evolutionary process is to lexicalize novel ideas, therefore serving as an adaptive system which supports the symbolic encoding and communication of emerging concepts~\cite{deacon1997symbolic,pinker1990natural}. The most common strategies of lexicalization involve reusing and combining existing words in the lexicon~\cite{marchand1960wf,algeo1980all,brinton2005lexicalization,ramiro2018algorithms}, although other strategies such as borrowing (e.g., {\it tofu}) and coinage (e.g., {\it quark}) also exist. Reuse refers to using the form of an existing word to express something new. 
For instance, {\it mouse} was reused to describe a ``small device that is moved by hand across a surface to control the movement of the cursor on a computer screen''~\cite{oedmouse}. Combination refers to concatenating two or more existing words to form a new word, typically known as a compound (e.g., {\it armchair} combines {\it arm} and {\it chair} to express a new type of chair). Word reuse and combination are often viewed and studied separately as two distinct aspects of lexical evolution. Here we present an information-theoretic framework that accounts for both strategies through the lens of efficient communication.

Word reuse has been traditionally discussed in the context of historical semantic change~\cite<e.g.,>{traugott_dasher_2001} and word meaning extension~\cite<e.g.,>{williams1976synaesthetic}. This body of work aims to identify regularities in how words take on new meanings over time. More recent studies using large-scale historical and cross-linguistic data have suggested that words tend to take on new meanings that are semantically related to existing ones~\cite{xu2016historical,ramiro2018algorithms}, and the processes of word meaning extension reflect cognitively economic ways of expanding the referential range of existing words in the lexicon~\cite{srinivasan2015concepts,xu2020conceptual}. However, this line of research focuses almost exclusively on word reuse (i.e., with no overt changes in word form) and treats it in isolation from word combination.

% Work on combinations, no existing account for both reuse and combination
Word combination has been commonly studied in the literature on word formation and morphology~\cite<e.g.,>{vstekauer2006handbook}. In particular, existing accounts focusing on noun-noun compounds have suggested that compound interpretation involves selecting from a systematic list of predicate relations which in turn constrains possible noun combinations produced by speakers~\cite<e.g.,>{levi1978compound,lieber1983argument,levin2019systematicity}. An alternative approach appeals to more functionally motivated principles arguing that compounds should be semantically transparent while shorter forms are preferred if compound constituents are redundant~\cite{downing1977creation,dressler2005word,costello2000efficient}. These principles have also been discussed in psycholinguistic work showing that the semantic relatedness between a novel compound and its constituents predicts its acceptability, unless the relatedness is too high~\cite{gunther2016understanding}. We believe these functional principles might be equally applicable to explaining word reuse. However, to our knowledge there is no unified account that characterizes both word reuse and combination~\cite<c.f.>{blank2003words,xu2023strategy}.

We propose a unified account of word reuse and combination by building on the view that language is shaped to support efficient communication~\cite<e.g.,>{regier2015word,kemp2018semantic,gibson2019efficiency,hahn2020universals}. This line of work suggests that linguistic structures are shaped by functional pressures to maximize informativeness and simplicity (or ease of use) in communication. Efficiency-based accounts have been shown to explain word meaning variation across languages~\cite<e.g.,>{kemp2012kinship,regier2015word,xu2020numeral,zaslavsky2018efficient}, the structures of word forms~\cite<e.g.,>{zipf1949human,mahowald2018word,bentz2016zipf,hahn-2022-morpheme}, and grammatical form-meaning mappings~\cite{mollica2021forms}. We extend this growing body of research with the aim to understand the general principles that shape the diverse strategies of lexicalization.

\begin{figure*}[ht!]
\centering
\includegraphics[width=\linewidth]{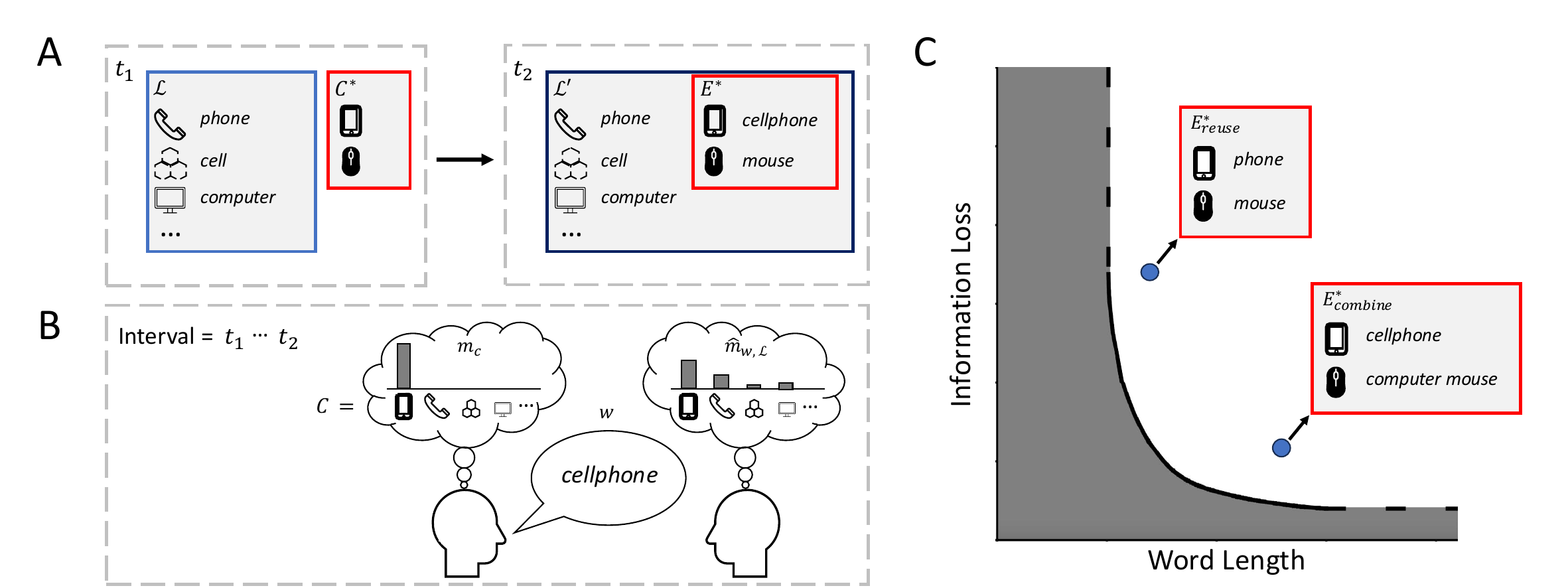}
\caption{Illustration of our theoretical proposal. Panel (A) illustrates the lexicalization of emerging concepts using examples from English during the historical interval 1980-2000. The existing lexicon $\mathcal{L}$ and the set of emerging concepts $\mathcal{C}^*$ at time $t_1$ are illustrated on the left. At a later time $t_2$, the attested encoding of the novel concepts $\encoding$ enters the expanded lexicon $\mathcal{L}'$, which are shown on the right. Panel (B) illustrates the two opposing pressures in a communicative interaction taking place before $t_2$. Here the speaker intends to convey the emerging concept ``cellphone'' to a listener whose lexicon does not yet have a word for expressing it, and grey bars illustrate probability distributions over a universe of concepts $\mathcal{C}$ that capture uncertainty regarding the intended concept. Our proposal focuses on the pressure for minimizing the length of the utterance, and the pressure for minimizing information loss, or the difference between the speaker and listener distributions over concepts. Panel (C) illustrates possible encodings of the novel concepts in Panel (A). Each point corresponds to the average length and information loss of an encoding of the novel concepts, and the shaded area corresponds to costs that are not attainable. We propose that word reuse and combination reflect a tradeoff between these two costs, and that both attested reuse items and attested compounds achieve tradeoffs that are relatively efficient. Here the example encodings are simplified to contrast reuse and combination, and in reality an encoding can consist of both strategies.
%such that the costs it incurs interpolate between the extremes.
}\label{fig:intro}
\end{figure*}

Here we extend existing efficiency-based accounts that assume speakers and listeners use the same static lexicon by considering communicative interactions during the spread of linguistic innovations~\cite<e.g.,>{labov2011principles,milroy1985linguistic}. In particular, we consider the labels of novel concepts that are yet to be encoded in the lexicon of a large number of language users. We propose that when speakers communicate novel concepts to these language users, word reuse and combination reflect a fundamental tradeoff between speaker effort and information loss (or the inverse of informativeness): on the one hand, speakers can minimize their effort by reusing short words that underspecify intended concepts; on the other hand, information loss is minimized if speakers use relatively long word forms that combine existing words in an informative way. We hypothesize that as speakers repeat these communicative interactions, their encoding of novel concepts is shaped by the pressure to minimize speaker effort and the opposing pressure to minimize information loss, such that the word length and informativeness of both attested reuse items and attested compounds efficiently trade off against each other. We illustrate this idea in Figure~\ref{fig:intro}.
% "emerging" is better in context of t1, "novel" is better in context of listener/communication

% ($\encoding_{\text{reuse}}$ and $\encoding_{\text{combine}}$)
% Our theory predicts that word reuse and combination enable a tradeoff between these two costs, and that both attested reuse items and attested compounds achieve tradeoffs that are relatively efficient.

% General relations with other functionalist work
Our proposal is consistent with other functional accounts of word reuse and combination. Previous accounts have separately suggested that combinations should be informative~\cite{lieber_2004,clark1984structure,downing1977creation} and reuse represents an economical strategy~\cite{vstekauer2005onomasiological,piantadosi2012communicative}. Computational studies of reuse items~\cite{ramiro2018algorithms,brochhagen2023language} and compounds~\cite{gunther2016understanding,vecchi2017spicy,pugacheva2024lexical} have shown that semantic transparency is preferred across both strategies, which is consistent with a preference for using informative word forms to reduce communicative error. In recent work on language evolution, a pressure for informativeness is often considered necessary for compositional or subword structure to emerge in the lexicon~\cite<e.g.,>{kirby2015compression,carr2017cultural}. Crucially, prior studies have shown that a pressure for informativeness may be sufficient for the emergence of subword structure when speakers have to communicate novel meanings to a large community with limited shared history~\cite{raviv2019compositional,raviv2019larger}. Here we build on these existing studies to offer a unified functional account of word reuse and combination.

% Outline of remainder of paper
In the following, we first specify our theoretical proposal in formal terms. We then evaluate our efficiency-based proposal against a large historical dataset of reuse items and compounds attested in English, French, and Finnish over the past century. Lastly, we discuss the implications of our results and avenues for future work.

\section{Computational Formulation of Theory}
To specify our theoretical proposal, we formulate a scenario in which the attested encoding of emerging concepts spreads within a speech community. Our formulation builds on a standard approach in language change~\cite<e.g.,>{weinreich1968empirical,milroy1985linguistic,traugott_dasher_2001,brinton2005lexicalization,labov2011principles}, which views the evolution of linguistic structures as a gradual process in which new structures coexist with existing structures as the former spread among speakers. 

We illustrate the setting for this scenario in Figure~\ref{fig:intro}A. Here we consider an encoding of concepts as a set of form-concept pairs (or mappings), and we treat the lexicon as an encoding of lexicalized concepts which may incorporate novel pairs over time. Suppose the existing lexicon $\mathcal{L}$ consists of form-concept pairs known by all speakers in the speech community at time $t_1$, and the set $\mathcal{C}^*$ contains novel concepts emerging at $t_1$ but not encoded in the existing lexicon. We assume that the eventual attested encoding of novel concepts, denoted by $\encoding$, spreads among speakers until the expanded lexicon $\mathcal{L}' = \mathcal{L} \cup \encoding$ has been acquired by all speakers at time $t_2$. In the current study, we assume that forms in $\encoding$ always reuse or combine forms that exist in $\mathcal{L}$.
% but this assumption may be lifted to account for other types of word formation in future work. 

In the following, we will focus on the time interval between $t_1$ and $t_2$ in which both the existing and expanded lexicons coexist within the speech community. We first specify an information-theoretic model of communication. We then build on the model to define two types of communicative cost and specify our theoretical proposal regarding forms in $\encoding$.
% $I \subseteq [t_1, t_2]$
% I = [t_1, t], t \le t_2

\subsection{Model of Communication} To assess the role of communicative efficiency in shaping $\encoding$, we first consider the communicative interaction between a speaker who uses the expanded lexicon $\mathcal{L}'$ and a listener who uses the existing lexicon $\mathcal{L}$. We model this interaction by extending previous efficiency-based accounts of the lexicon~\cite{kemp2018semantic,zaslavsky2018efficient} that are grounded in Shannon's original point-to-point model of communication~\cite{shannon1948mathematical}.
% and formulate information loss in a way related to the information bottleneck~\cite{tishby2000information}. 
% We further assume $\mathcal{C}^* \subset \mathcal{C}$, i.e., emerging concepts are also plausible.

We describe our model using an illustration of the interaction in Figure~\ref{fig:intro}B. The speaker's mental representation is a speaker distribution $m_c$ over a universe of concepts $\mathcal{C}$. % that contains concepts that the speaker and listener find plausible or consistent with prior knowledge~\cite{costello2000efficient}. 
In general $m_c$ could capture speaker uncertainty, but we assume that $m_c$ corresponds to a single intended concept $c$ and picks it out with certainty. The intended concept $c$ is drawn from a need distribution $p(c | \mathcal{L}')$ which captures the frequency with which different concepts are communicated~\cite<e.g.,>{kemp2018semantic,zaslavsky2018efficient}, and here we assume the speaker only communicates concepts encoded in her lexicon. To express $c$, the speaker selects a form $w$ according to her production policy $p(w|c, \mathcal{L}')$ which captures the frequency of using specific forms in her lexicon to communicate an intended concept. In turn, the listener uses $w$ and his lexicon to deterministically construct his mental representation which is a listener distribution $\hat{m}_{w,\mathcal{L}}$ that aims to reconstruct $m_c$. 

We define the listener distribution by using a variant of prototype-based categorization models~\cite<e.g.,>{rosch1975cognitive,ramiro2018algorithms}. In our model, the listener treats each form $w$ as the label of a category of concepts, which is represented by the category prototype $\prototype_{w}$. The listener uses the category to construct a distribution, so that the probability of concept $c$ is high if it is semantically similar to $\prototype_{w}$. We specify this distribution via the similarity choice model~\cite{luce1963detection,nosofsky1986attention}:
\begin{align}
    \hat{m}_{w,\mathcal{L}}(c) \propto \exp{\left\{-\gamma d(c, \prototype_{w})\right\}}
    % &= \text{sim}(c, \prototype_{w,\mathcal{L}}) \\
    \label{eq:sim_func}
\end{align}
where $d(\cdot, \cdot)$ is semantic distance, and $\gamma \ge 0$ is a sensitivity parameter that controls how fast probability decreases with distance. We require the prototype to be a function of form-concept pairs in $\mathcal{L}$, but its exact definition depends on the specific dataset to which we apply our framework.
% this ensures the listener distribution is independent of how novel concepts are encoded in $\mathcal{L}'$.
% Equation~\ref{eq:sim_func}
% Following previous models of prototypes (e.g., refs.~\citeA{reed1972pattern,smith1988combining}), $\prototype_{w}$ is a function of concepts that are paired with $w$ or the constituents of $w$ in lexicon $\mathcal{L}$.
% that controls how fast probability decreases with distance. 
%and it has no direct access to how novel concepts are encoded in the speaker's lexicon $\mathcal{L}'$.
% TODO: probably want to say this can be related to various computational accounts in lexical semantics, that are related to both reuse and compound

\subsection{Communicative Costs} 
To specify our theoretical proposal, we now consider the speaker effort and information loss incurred over repetitions of the above interaction. In reality, these interactions take place across multiple speakers and repeatedly within the same dyads, but these dynamics introduce heterogeneity among listener distributions as listeners adopt new form-concept pairs into their lexicons. For simplicity, we consider the case in which a single speaker interacts once with each of many distinct listeners, such that the listener distributions are independent and identical across interactions. The speaker in this special case may be construed as a leader in spreading linguistic innovations in local communities~\cite<e.g.,>{labov2011principles,milroy1985linguistic,del2018road}.
% This special case is related to sociolinguistic accounts of language change, which have shown that well-connected community leaders~\cite{labov1972language} or innovators with ties across many communities~\cite{milroy1985linguistic} are central to the spread of linguistic innovations.

%The first communicative cost relevant to our proposal is speaker effort. 
Following efficiency-based accounts of word length~\cite<e.g.,>{zipf1949human,mollica2021forms}, we measure the speaker effort in an interaction via the length of the produced utterance. As the same speaker communicates with many listeners, the average speaker effort over their interactions is given by expected word length:
\begin{equation}
    \mathbb{E}[l(W) | \mathcal{L}'] = \sum_{c, w}p(c, w| \mathcal{L}')l(w)
    \label{eq:avg_length}
\end{equation}
where $l(\cdot)$ is the length of a form. Previous studies on coding efficiency have shown that word frequency and length tend to be related in a way that is relatively efficient~\cite<e.g.,>{zipf1949human,bentz2016zipf,mollica2021forms}. Here we extend these studies by examining the tradeoff between length and information loss in reused and compound forms that express novel concepts.

% and $p(c, w| \mathcal{L}')$ is the product of the need distribution and the speaker's production policy.

% The second communicative cost relevant to our proposal is the listener error in reconstructing the speaker distribution. 
% refs.~\citeA{regier2015word} and~\citeA{zaslavsky2018efficient}
Following~\citeA{regier2015word}, we define the information loss in a single interaction as the Kullback–Leibler (KL) divergence between the speaker and listener distributions. In our case of a single speaker and many distinct listeners, the average information loss over their interactions is given by expected KL divergence:
\begin{align}
    \mathbb{E}[D(M || \hat{M}) | \mathcal{L}', \mathcal{L}] = \sum_{c,w}p(c, w| \mathcal{L}') h(\hat{m}_{w,\mathcal{L}}(c))
    \label{eq:distortion2}
\end{align}
where $h(\cdot) = - \log_2{(\cdot)}$. In contrast to previous efficiency-based accounts of lexical semantics~\cite<e.g.,>{regier2015word,zaslavsky2018efficient}, here we consider information loss when the production policy and the listener distribution are conditioned on different lexicons.
% TDODO: make this last sent more informative

% \begin{align}
%     D(m_c || \hat{m}_{w,\mathcal{L}}) &= \sum_{u\in \mathcal{C}} m_{c}(u)\log_2{\frac{m_{c}(u)}{\hat{m}_{w,\mathcal{L}}(u)}} =  h(\hat{m}_{w,\mathcal{L}}(c))
%     \label{eq:info_loss}
% \end{align}
% The last term reduces divergence to surprisal, which measures how surprised each listener should be to uncover the intended concept.
 
% If the form $w$ is ambiguous, the listener will generally not be able to approximate the speaker distribution with perfect accuracy. 
% where divergence reduces to surprisal due to Equation~\ref{eq:info_loss}. 
% The last step follows from the assumption that the speaker mentally represents her intended concept with perfect accuracy. % i.e., $m_{c}(u) = 1$ if $c=u$. 

\subsection{Efficiency of Attested Encodings}
Our proposal can be specified by considering how much the attested encoding $\encoding$ contributes to these communicative costs. This contribution can be summarized by a single objective function obtained from combining and simplifying Equations~\ref{eq:avg_length} and~\ref{eq:distortion2}:
\begin{align}
    L_\beta[\encoding | \mathcal{L}] &= \mathbb{E}[D(M || \hat{M}) | \mathcal{L}', \mathcal{L}] + \beta\mathbb{E}[l(W) | \mathcal{L}'] \\
    &\propto \sum_{ (c, w) \in \encoding} p(c, w| \mathcal{L}') \cdot \left( h(\hat{m}_{w,\mathcal{L}}(c))  + \beta l(w) \right)
    \label{eq:objective-rewrite}
\end{align}
where $\beta \ge 0$ is a tradeoff parameter. As in previous efficiency-based approaches~\cite{zaslavsky2018efficient,mollica2021forms}, optimizing Equation~\ref{eq:objective-rewrite} for every $\beta$ produces a Pareto frontier that specifies the space of possible encodings of $\mathcal{C}^*$, which we illustrate in Figure~\ref{fig:intro}C. We hypothesize that the attested encoding is shaped by competing pressures for minimizing speaker effort and information loss, which predicts that it will be relatively close to the Pareto frontier in this space of possible encodings.
%  that determines the overall importance of minimizing speaker effort over informativeness

% \begin{equation} \label{eq:objective-rewrite}
% \begin{split}
%     L_\beta[\encoding | \mathcal{L}] &= \mathbb{E}[D(M || \hat{M}) | \mathcal{L}', \mathcal{L}] + \beta\mathbb{E}[l(W) | \mathcal{L}'] \\
%     &\propto \sum_{ (c, w) \in \encoding} p(c, w| \mathcal{L}') \cdot \left( h(\hat{m}_{w,\mathcal{L}}(c))  + \beta l(w) \right)
% \end{split}
% \end{equation}

In {\it SI Appendix, Section S1}, we provide a more detailed derivation of Equations~\ref{eq:avg_length}-\ref{eq:objective-rewrite}. We also provide a discussion on the limitations of previous efficiency-based accounts of the lexicon in terms of accounting for attested combinations of words or morphemes.

% % TODO: make a figure to illusrate pipeline

\section{Results}
To test our proposal, we instantiated our scenario using Princeton WordNet~\cite{fellbaum1998wordnet} and its multilingual extensions~\cite{bond2013linking}. These WordNets are conceptually organized dictionaries that map language-specific, orthographic forms to a common set of word senses or lexicalized concepts. We focused on English, French, and Finnish because they have the largest WordNets among alphabetical languages in terms of the size of their sense inventory~\cite{bond2013linking}. For each language and one of five consecutive intervals over the past century, we instantiated emerging concepts as WordNet senses, and we instantiated their attested encoding and the existing lexicon as sets of form-sense pairs. We identified whether an English form-sense pair is an emerging reuse item or compound by using its first citation in the Historical Thesaurus of English~\cite{HTE4.2}, and we inferred whether a pair is existing based on its estimated frequency in historical text~\cite{davies2002corpus,michel2011quantitative}. We implemented the same components for French and Finnish by relabelling emerging and existing senses in the English data with a language-specific form that was attested in historical French or Finnish text~\cite{michel2011quantitative,FNC1_en}. For tractability, we ignored linking constituents in compounds and we used compounds that have exactly two constituents. More details on data processing are provided in {\it Materials and Methods}. 

% These components were based on first citations in the Historical Thesaurus of English~\cite{HTE4.2}, form-sense frequencies in historical corpora~\cite{davies2002corpus,michel2011quantitative,FNC1_en}, and the assumption that concepts emerge at approximately the same time across languages.
% we focused on five consecutive intervals over the twentieth century for tractability
%  $(\mathcal{C}, \mathcal{C}^*, \mathcal{L}, \encoding)$

% To show that our approach applies to both reuse and combination, 
To show that our approach applies to both reuse and combination, we controlled for differences in sample size between strategies by instantiating each attested encoding such that it only contains reuse items or compounds. Across our target intervals, we analyzed $518$ reuse items and $2,828$ compounds in English, $529$ reuse items and $409$ compounds in French, and $510$ reuse items and $645$ compounds in Finnish; sample sizes for specific intervals are provided in {\it SI Appendix, Section S2}. In Table~\ref{table:data_examples}, we show examples of reuse items and compounds that make up these encodings.

\begin{table*}[ht!]
\small
\centering
\begin{tabular}{lllll} 

{\bf Language} & {\bf Interval} & {\bf Strategy} & {\bf Form} &  {\bf Sense Definition} \\
\hline
English & $1940+$ & R & locker & a trunk for storing personal ... \\

  & $1940+$ & R & printer &  an output device that prints the ... \\

 & $1940+$ & R & dish &  directional antenna consisting of a ... \\
 & $1900+$ & C & birthday card &  a card expressing a birthday ... \\

 & $1940+$ & C & urban renewal & the clearing and rebuilding and ... \\

 & $1980+$ & C & spreadsheet &  a screen-oriented interactive ... \\
%\hline
French & $1900+$ & R &  antenne & an electrical device that sends or ...\\

 & $1920+$ & R & publicité &  a commercially sponsored ad on  ... \\

 & $1960+$ & R & émuler & imitate the function of ... \\

 & $1900+$ & C & turbine à gaz & turbine that converts the chemical  ... \\

  & $1900+$ & C & galaxie spirale & a galaxy having a spiral structure; ... \\

  & $1940+$ & C & boîte noire &  equipment that records information ... \\
%\hline
Finnish & $1900+$ & R & lähetys & message that is transmitted by ... \\

 & $1920+$ & R & suodatin & an air filter on the end of a ... \\

 & $1940+$ & R & ajaa & carry out a process or program, as  ... \\

  & $1900+$ & C & sotarikos &  a crime committed in wartime; ... \\
  
  & $1900+$ & C &  taisteluväsymys & a mental disorder caused by stress ... \\

  & $1920+$  & C & kauppa-apulainen & a salesperson in a store \\

\end{tabular}
\caption{Examples of reuse items (R) and compounds (C) that emerged in the past century; sense definitions have been truncated for brevity}
\label{table:data_examples}
\end{table*}

Given the existing lexicon, we computed the average length and information loss incurred by a speaker communicating with an encoding of novel concepts as specified in Equation~\ref{eq:objective-rewrite}. We used the orthographic length of word forms in our subsequent analyses as a proxy for production effort. To estimate information loss, we implemented the listener distribution in Equation~\ref{eq:sim_func} in three parts. We first represented each concept or word sense by embedding the text of its WordNet definition (see Table~\ref{table:data_examples} for examples) with a sentence encoder~\cite{reimers2019sentence}, and we followed approaches that construct the prototype as an average of the existing senses of a word or its constituents~\cite{reed1972pattern,mitchell-lapata-2008-vector}. For the sensitivity parameter, we set $\gamma = 10$ based on the informativeness of existing words. These costs were computed for each form-sense pair in the encoding, and then averaged according to their need and production probabilities estimated from historical form-sense frequencies. We specify our implementation in {\it Materials and Methods}.
% and its justification

In the following, we first directly assess the average-case efficiency of attested reuse-based and combination-based encodings. We then perform fine-grained analyses on the efficiency of individual reuse items and compounds.

\subsection{Average-case Efficiency} In our first analysis, we compared attested items to optimal encodings on the Pareto frontier and two sets of baseline encodings. The first baseline consists of alternate encodings created from replacing the label of each item in an attested encoding with a near-synonym; examples of near-synonyms are shown in Table~\ref{table:ns_examples}. To probe the space of all possible alternatives, we created a second baseline by replacing each attested label with a string uniformly sampled from labels in the existing lexicon and their combinations. Details on estimating the Pareto frontier and creation of baseline encodings are specified in {\it Materials and Methods}.
% $\encoding'$

\begin{table}[t!]
\centering
\begin{tabular}{ll} 
%\hline
{\bf Attested Label} & {\bf Near-Synonyms} \\
\hline
locker & deedbox, strongbox, clothespress, storeroom \\ % locker \\
urban renewal & renewal, renovation, urban-renovation \\ %  \\

publicité & réclame, annonce, pub, emballage \\ % publicité
turbine à gaz & turbine, générateur, turbine-fluide, moteur-gaz \\ % générateur-gaz \\

%suodatin & pyyhin, laskuputki, puhdistusväline, asetin \\ 
 lähetys & lasti, rahti, toimitus, kuorma \\
%\hline
sotarikos & rikos, laittomuus, sota-laittomuus, hyökätä-rikos \\ % lainvastaisuus \\
%\hline
\end{tabular}
\caption{Samples of near-synonyms created for attested labels}
\label{table:ns_examples}
\end{table}

\begin{figure*}[!ht]
\centering
% \includesvg[width=\linewidth]{figures/average_comparisons_wn.svg}
\includegraphics[width=\linewidth]{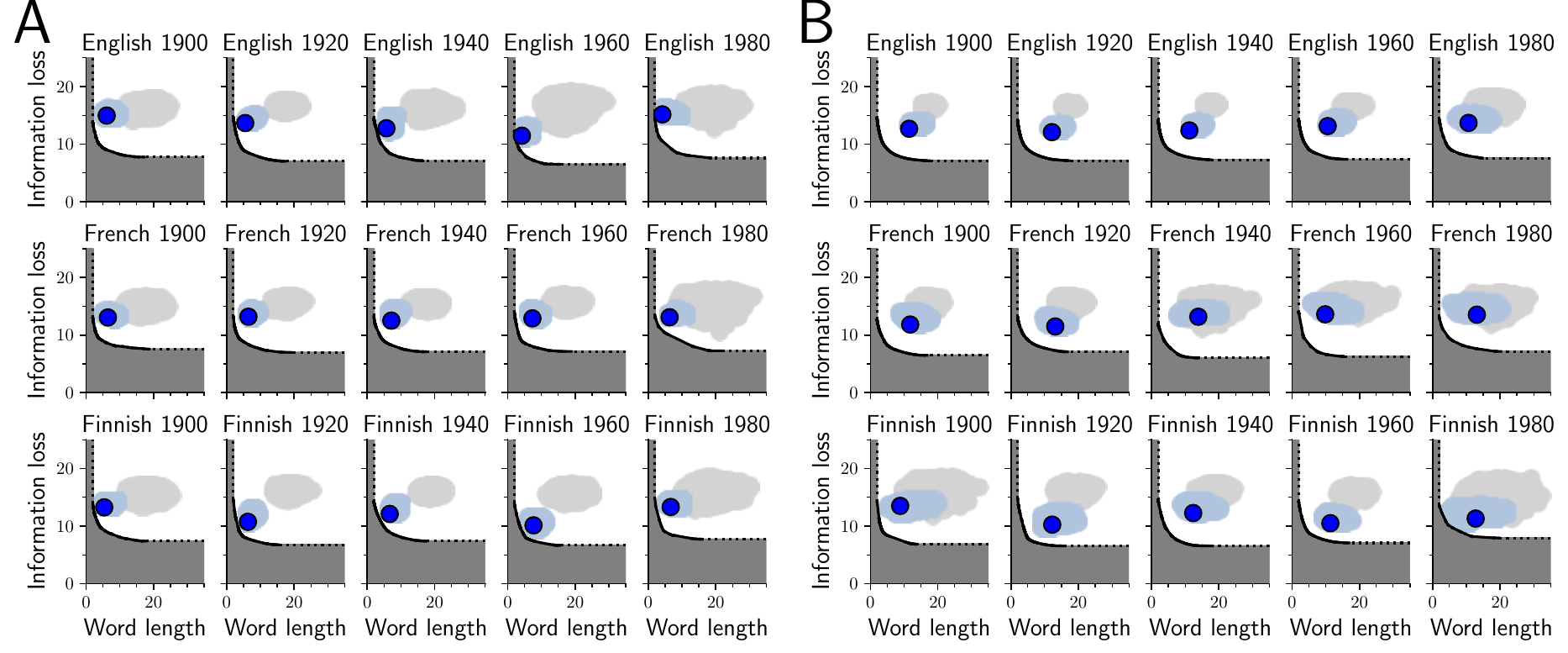}
\caption{Illustration comparing (A) attested reuse items and (B) attested compounds to the constructed baselines and the Pareto frontier. Every point corresponds to an encoding of emerging concepts for a specific language and interval. Attested cases are marked in blue, near-synonym baselines in light blue, and random baselines in grey. Black solid lines in the bottom left show the estimated Pareto frontier, and the shaded areas show costs that are not attainable.
}
\label{fig:average_summary}
\end{figure*}
% the size of markers for attested cases is larger than the size of other markers for improved visibility.
% Note that the axes have different scales, and the x-axis increases faster in magnitude over the same visual distance than the y-axis.

\begin{figure*}[!ht]
\centering
% \includesvg[width=1\linewidth]{figures/average_epsilon_wn.svg}
\includegraphics[width=0.7\linewidth]{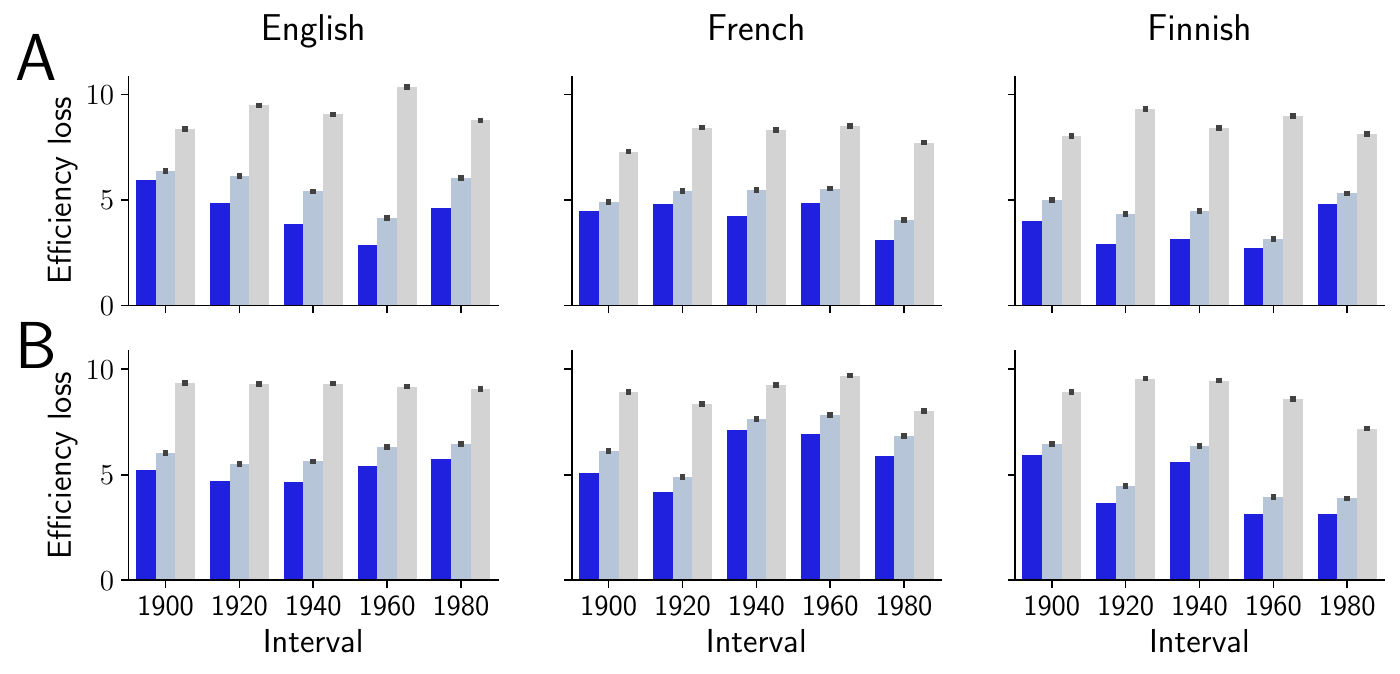}
\caption{Efficiency loss of attested encodings for (A) reuse items and (B) compounds relative to the average loss of baselines. Attested loss is marked in blue, and the average loss of near-synonym and random baselines is marked in light blue and grey, respectively. Error bars show bootstrapped 95\% confidence intervals.
}
\label{fig:average_loss}
\end{figure*}

Figure~\ref{fig:average_summary} summarizes the comparisons between attested and alternative encodings. Each Pareto frontier shows the optimal tradeoff that can be achieved by any encoding of the emerging concepts, and intuitively, the closeness of an encoding to the frontier approximates its efficiency. Across strategies, intervals and languages, we observe that both attested encodings (blue) and near-synonym baselines (light blue) tend to be closer to the frontier and more efficient than random baselines (grey), and attested encodings tend to be more efficient than both baselines. By construction, attested and near-synonym encodings tend to be shorter than random baselines because it is more likely to sample a combination of long words than a single short word. The fact that attested labels also dominate near-synonyms indicates the relative efficiency of attested labels does not arise solely due to chance and the prevalence of longer word forms.
% $\mathcal{C}^*$
%or by statistical properties of word lengths.

Figure~\ref{fig:average_loss} compares attested encodings against baselines using a quantitative measure of efficiency loss (see {\it Materials and Methods}), which overall confirms our qualitative observations. In {\it SI Appendix, Section S4}, we show that attested reuse items and compounds remain more efficient than the constructed baselines under different implementations of our scenario of lexical evolution. First, we show our results are robust if we use a uniform distribution over attested items and different values for the sensitivity parameter. Second, we show the results hold up across different communication channels via an implementation that represents word forms using phonemes instead of letters. Third, we describe an analysis based on historical embeddings~\cite{mikolov2013efficient,hamilton-etal-2016-diachronic} to address the concern that our approach may be biased by using contemporary embeddings to study historical change. Lastly, we show these findings are robust to alternative datasets of lexicalized concepts by considering another English dictionary and by assuming one-to-one correspondence between concept and form.

\subsection{Item-level Variation} In Figure~\ref{fig:average_summary}, we observe non-trivial gaps between Pareto frontiers and attested encodings. This suggests that the communicative efficiency of attested encodings could be improved by replacing some attested reuse items and compounds with more efficient forms. We thus compared individual items to optimized forms by using the same implementation of our scenario, except we replaced full encodings with singletons that contain individual items. 

\begin{figure*}[!t]
\centering
%\includesvg[width=1\linewidth]{figures/item_eff_loss.svg}
\includegraphics[width=0.7\linewidth]{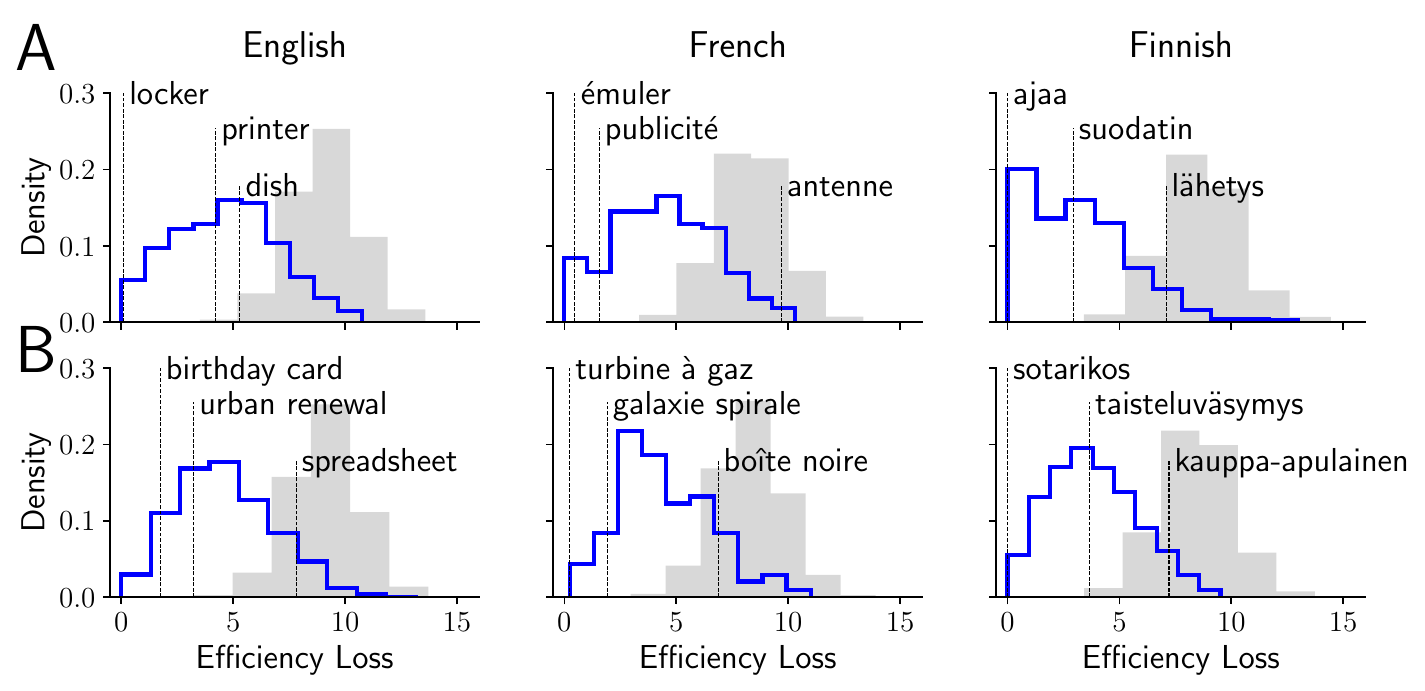}
\caption{Efficiency loss of individual attested items for (A) reuse and (B) compounding and randomly sampled labels. The distributions for attested and random are marked in blue and grey, respectively. Examples in Table~\ref{table:data_examples} are annotated.
}
\label{fig:item_loss}
\end{figure*}

\begin{figure*}[!ht]
\centering
% \includesvg[width=\linewidth]{figures/item_level_small.svg}
\includegraphics[width=\linewidth]{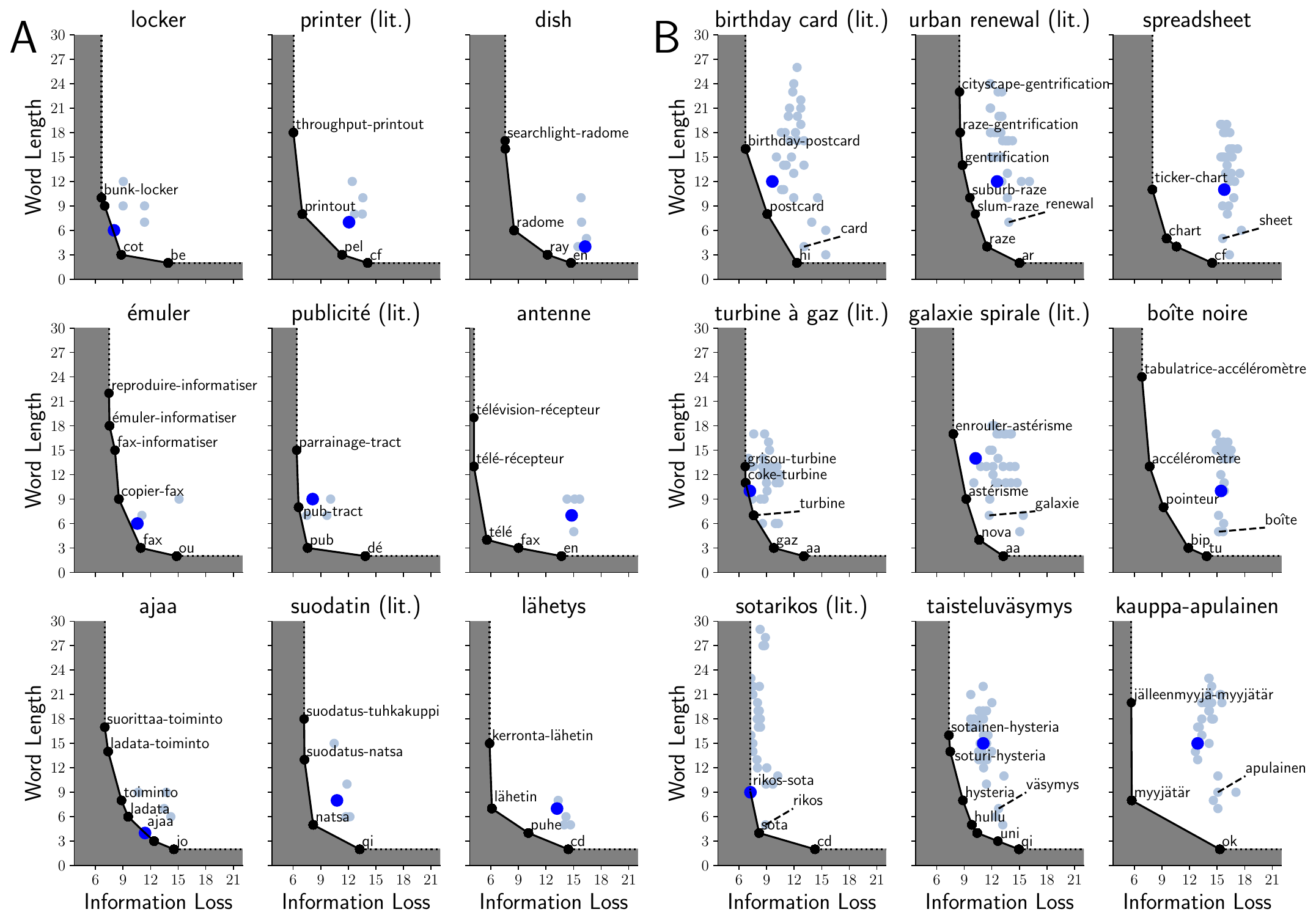}
\caption{
Item-level illustration for (A) attested reuse items and (B) attested compounds. Headers correspond to the examples in Table~\ref{table:data_examples}, with additional marking for literal items (lit.). Each dark blue dot corresponds to an attested form. Black dots correspond to the item-level Pareto frontier, and light blue dots correspond to the near-synonym set generated for this item; the size of markers for attested items is larger than the size of other markers for improved visibility. A sample of optimal labels and compound head words are shown as text. Note that the axes are swapped relative to Figure~\ref{fig:average_summary} and the x-axis is truncated so there is more space to display optimal labels. }
\label{fig:item_examples}
\end{figure*}

Figure~\ref{fig:item_loss} shows efficiency losses for individual items, which measure their deviation from optimized forms, and each distribution is aggregated over all time intervals. As in the previous analysis, the item-level loss is approximated by the distance between attested items and Pareto frontiers, which is illustrated in Figure~\ref{fig:item_examples}. {We observe that attested items tend to be much closer to optimized forms (loss = 0) than most randomly sampled labels (grey), but nonetheless the right tails of attested items overlap with random distributions}. In {\it SI Appendix, Section S5.A}, we show that item-level loss based on orthographic forms strongly correlates with item-level loss based on phonemic forms. The variation from near-optimal to near-random among attested reuse items and compounds reveals that some items are more strongly shaped by our proposed tradeoff than others. 

% This suggests the efficiency of attested encodings could be improved by replacing some of its labels with more efficient forms. To examine how individual items support or reduce communicative efficiency, we examined the tradeoff between length and information loss at the item level: for each attested item or form-sense pair, we computed its efficiency loss using the same quantitative measure, except we replaced the full encoding with a singleton containing the item. Figure~\ref{fig:item_loss} shows the efficiency loss of items aggregated over all time intervals for English, French, and Finnish. As in the average-case analysis, the item-level loss is approximated by the distance attested items are from the frontier. This is illustrated in Figure~\ref{fig:item_examples} for the examples in Table~\ref{table:data_examples}. Here we observe distributions of item-level loss tend to have wide spread that is also right-skewed. In {\it SI Appendix, Section S6.A}, we show that item-level loss based on orthographic forms strongly correlates with item-level loss based on phonological forms. {\color{blue} TBC}
% with a shorter form of equal information loss or an equally long form that is more informative.

Here we characterize this variation using two well-studied subclasses of lexical items. Endocentric compounds are the most well-studied subclass of compounds that are defined by the relation between intended and existing constituent concepts~\cite<e.g.,>{downing1977creation,jackendoff2010ecology}. The head word of an endocentric compound encodes a superordinate category of the intended concept and is a literal expression of this concept (e.g., {\it birthday card} is a {\it card}), in contrast to non-literal (or exocentric) compounds (e.g., {\it blue-collar}). Similarly, a subclass of reuse items often expresses an intended concept that is more narrow than an existing sense of the reused word~\cite<e.g.,>{bloomfield1933language}, for instance the modern use of {\it car} as a motorized vehicle refers to a narrower set of concepts compared to its original meaning of {\it wheeled cart}. Across strategies, these literal expressions may be more efficient because they are more transparent to the listener than non-literal ones. For example, in Figure~\ref{fig:item_examples}, we observe that literal reuse items and endocentric compounds tend to be more efficient than their non-literal counterparts (e.g., {\it birthday card} vs {\it dish} or {\it antenne}).

To test this hypothesis, we leveraged the WordNet taxonomic hierarchy to classify reuse items and compounds into literal and non-literal cases. We performed a quantitative analysis that compares the efficiency loss of literal and non-literal items; we supplemented attested reuse items with additional data by using the head words of attested compounds, since the original English WordNet does not explicitly encode literal items~\cite{miller1998nouns} (see {\it Materials and Methods} for details). We find that in French and Finnish, literal reuse items are significantly more efficient than non-literal items ($t(527) = 4.70$, $p < .001$; $t(508) = 6.37$, $p < .001$), and the same trend holds between literal and non-literal head words in English ($t(2793) = 23.60$, $p < .001$), French ($t(396) = 7.32$, $p < .001$), and Finnish ($t(631) = 9.40$, $p < .001$); endocentric compounds also tend to be more efficient on average in English ($t(2826) = 17.26$, $p < .001$), French ($t(407) = 3.68$, $p < .001$), and Finnish ($t(643) = 7.58$, $p < .001$). We illustrate these comparisons in {\it SI Appendix, Section S5.D}. These results suggest our efficient tradeoff proposal applies more strongly to labels that encode novel concepts in a literal way across both strategies.
% non-literal reuse vs endocentric head: English ($t(1649) = 19.81$, $p < .001$), Finnish ($t(745) = 8.28$, $p < .001$), and French ($t(666) = 9.61$, $p < .001$)

In {\it SI Appendix, Section S5.E and Section S5.F}, we explore the variation in efficiency among attested reuse items and compounds in two further analyses. In the first analysis, we used taxonomic distance measures~\cite{wu1994verbs,leacock1998using} as a continuous version of the literal and non-literal distinction. In line with our findings above, we found that taxonomic distance measures are positively correlated with efficiency loss across all languages and across both strategies. In the second analysis, we investigated whether frequent items are closer to Pareto frontiers since this implies less total efficiency loss. We did not find that frequency differentiates variation in item-level loss. This may be due to frequency effects in lexicalization beyond the scope of our account. We return to other factors that underlie lexical evolution in {\it Discussion}.

% For example, frequent items may reflect lexical strength~\cite{bybee1998emergent} and tend to stay in the lexicon regardless of communicative efficiency as defined in our proposal.
%Future work may investigate how frequency interacts with our account.

We demonstrate our findings with examples in Figure~\ref{fig:item_examples}. Along each Pareto frontier, optimal labels gradually increase in length from the shortest but uninformative words (e.g., {\it be}) to the most informative compounds. We observe that near-optimal items are qualitatively similar to these optimal labels (e.g., {\it locker} and {\it bunk locker}; {\it birthday card} and {\it birthday postcard}). On the other hand, suboptimal items like {\it dish} and {\it spreadsheet} tend to relate to the intended concept in a less literal way when compared to optimal labels. These examples showcase the finding that both attested reuse items and combinations are in part explained by an efficient tradeoff between word length and informativeness.

\subsection{Strategy Comparison} In Figure~\ref{fig:average_summary}, we also observe that reuse-based encodings and combination-based encodings tend to occupy different neighbourhoods in the space of possible encodings. To compare differences between the two strategies, we compared informativeness and word length between all attested reuse items and compounds, aggregated across intervals for each language. We show the statistics in {\it SI Appendix, Section S5.D}, finding that on average, attested reuse items tend to be shorter than attested compounds across all three languages, and attested compounds tend to be more informative than attested reuse items in English and French. This mirrors existing proposals that cast reuse as an economical lexicalization strategy~\cite{vstekauer2005onomasiological,piantadosi2012communicative} and compounding as an informative strategy~\cite{downing1977creation,clark1984structure}.

\section{Discussion}
% summarize findings
We have presented evidence that word reuse and combination are shaped by competing pressures of informativeness and length minimization that affect the lexicalization of emerging concepts. We formulated this view in information-theoretic terms, and we tested our proposal using large-scale resources over history and across languages. We found that both attested reuse items and attested compounds that emerged over the past century in English, French, and Finnish are more efficient than random and near-synonym baselines, and that literal items are generally more efficient than non-literal items across both strategies. %These results suggest word reuse and combination can be in part explained by our efficiency-based account.

% relate to EC
Our work makes several contributions to efficiency-based accounts of language. First, our work establishes a new connection between efficiency-based accounts and word formation. Previous accounts have focused on form length and meanings~\cite<e.g.,>{mollica2021forms} and morpheme ordering~\cite{hahn-2022-morpheme}, but not the specific combination of certain morphemes as opposed to others in compositional words. In particular, one account suggests that the presence of subword structures in word forms hinders efficient length minimization~\cite{pimentel-etal-2021-non}. We show that in the case of compounding, these structures may nonetheless support efficient communication in settings where the speaker and listener do not fully share the same lexicon. Second, another line of investigation uses ideas from information theory to show that existing form-meaning mappings support accurate reconstruction of intended meanings under cognitive constraints~\cite<e.g.,>{regier2015word,zaslavsky2018efficient}. Our work shows that information-theoretic formulations of efficiency can also account for generalizations to novel concepts not yet encoded in the lexicon. Lastly, previous work has also examined the tradeoff between simplicity and informativeness in the lexicon, but only within restricted domains~\cite{kemp2012kinship,xu2016historical,zaslavsky2018efficient,xu2020numeral,denic2020complexity,steinert2021quantifiers,zaslavsky2021let,chen2023information}. Our computational framework offers a promising venue for extending simplicity-informativeness analyses toward the broader lexicon beyond individual semantic domains. 
% Second, another line of investigation suggests that existing form-meaning mappings support accurate reconstruction of intended meanings under cognitive constraints (e.g., refs.~\citeA{regier2015word,zaslavsky2018efficient}).

% applications to morphology
Although we focused on the lexicalization strategies of reuse and compounding, the functional principles invoked by our theory are general and our framework can be applied to other types of word formation. For instance, derived words are also highly productive~\cite{algeo1980all} and previous work has suggested that they are subject to a preference for informativeness and brevity~\cite{lieber_2004,marelli2015affixation}. Since derived words are combinations of existing words and morphemes, one way to extend our approach might be to incorporate a representation of affix meanings and model their combination with stem words~\cite{marelli2015affixation,westbury2019conceptualizing}. Derivational morphemes tend to be shorter than free morphemes and might yield more efficient strategies for expressing emerging or novel concepts; for example, the derived word {\it physicist} is more compact than the compound {\it physics scientist}. As a result, derivation may sit at a midpoint between economy of production and informativeness, flanked by the strategies examined in the current study along the Pareto frontier of efficiency.

% discuss other factors: frequency, learning, social-cultural
% strategy choice; discuss length reduction
Our account suggests that speakers select for more efficient reuse items and compounds as they repeatedly communicate novel concepts to listeners, but it does not capture all aspects of the historical evolution of the lexicon. First, while both strategies can create efficient lexical labels, our account does not explain why certain concepts are encoded via compounding but not reuse or vice versa. Consistent with Zipf's law of abbreviation~\cite<e.g.,>{zipf1949human,bentz2016zipf,pimentel-etal-2021-non,kanwal2017zipf}, one possibility is that concepts with high communicative need are less likely to be expressed via compounds since they are relatively long and their morphological structure does not offer additional informativeness once they enter the listener's lexicon. Second, our account emphasizes the tradeoff between length and informativeness but there are complementary and potentially overriding factors. Recent work on language learning suggests new meanings are more easily learned if they are encoded by semantically transparent existing words~\cite{floyd2021children} or novel complex words~\cite{brusnighan2012combining}. Since informative reuse items and compounds also tend to be transparent, our results may be alternatively construed to reflect a cognitive pressure for ease of learning~\cite<c.f.>{brochhagen2022languages}. On the other hand, frequent lexical innovations are more likely to be picked up by speakers~\cite{bryden2018humans} and subsequently these lexical items are more likely to persist~\cite<e.g.,>{bybee1998emergent,pagel2007frequency,lieberman2007quantifying,hamilton-etal-2016-diachronic}. Frequent lexical innovations may outcompete alternative, more efficient innovations that express the same concepts but emerged later and were used less frequently by speakers.

% (e.g., refs.~\citeA{bybee1998emergent,pagel2007frequency,lieberman2007quantifying,hamilton-etal-2016-diachronic})

% Usage frequency is an important factor in lexical evolution, which often predicts frequent items are more resistant to change (e.g., refs.~\citeA{pagel2007frequency,lieberman2007quantifying,hamilton-etal-2016-diachronic}) and is likely because they are easier to access and more resistant to change once it enters the lexicon~\cite{bybee1998emergent}. Frequent lexical innovations are also more likely to be picked up by speakers (e.g., refs.~\citeA{bryden2018humans}). Innovations that are sufficiently frequent may enter the lexicon regardless of their communicative efficiency, which may be offset by other factors.

% For compounds, morphological family size is another predictor of labels for concepts. This may be related to anology-based processes that listeners employ to infer the meaning of a compound, but these processes are not captured by our implementation of the listener distribution.

% TODO: maybe add discussion on homogeneous vs heterogeneous speakers; previous ec work are all homogeneous; we make a first attempt at considering the latter setting

% limitations in data
Our work is limited in that meaning representations and concept emergence are derived from historical data based on English. Recent work suggests that word meanings across languages show considerable variation~\cite<e.g.,>{thompson2020cultural,lewis2023local} and our analysis of French and Finnish items may be revisited using representations entirely based on French and Finnish resources. However, adapting these representations to a historical setting is non-trivial. For example, unlike existing crosslinguistic studies on reuse that analyze word meanings on a per-word basis~\cite<e.g.,>{fugikawa2023computational}, our approach requires a large sample of existing form-sense pairs in a historical period. To apply our methodology for English to other languages, a large dataset of historical documents and high-quality historical dictionaries (e.g., COHA and the OED) are required, which is challenging in many languages. 

% An alternate approach would be to adapt our framework to non-historical exoteric settings (cf. ref.~\citeA{wray2007consequences}), but this offers a weaker test of our account since they take place after attested items have entered the lexicon.
% more directly derived from language users of the respective languages.

% extending the listener distribution
For simplicity, we measured the informativeness of reuse items or compounds by using representations that are derived from word sense definitions. This may have contributed to the result that literal reuse items and compounds are more efficient than non-literal cases, since the latter may reflect similarity and contiguity relations~\cite{bloomfield1933language,jackendoff2010ecology} that are not always encoded in sense definitions. For example, {\it computer memory} is a metaphorical extension of {\it human memory}, and the function of storage is encoded in both definitions; in contrast, visual similarities in the metaphor {\it computer mouse} and animal-related {\it mouse} are not directly encoded in their definitions. The situation for compounds is further complicated by the fact that the constituents may relate in a non-literal way (e.g., {\it ghost town}) or the constituents may relate in a literal way but their product may reflect non-literal extension (e.g., {\it white-collar}). Future work may build on recent work on polysemy and multi-modality~\cite{brochhagen2023language} and integrate it with computational models of compound interpretation~\cite<e.g.,>{mitchell-lapata-2008-vector,tratz2010taxonomy,nakov2013interpretation} to further differentiate genuinely inefficient labels (e.g., homonyms and opaque compounds) from other non-literal cases. 
% costello2000efficient

% summarize main contributions
In prior literature, word reuse and combination are typically treated as distinct areas of research. Our work provides a unified account of both lexicalization strategies by appealing to the general idea that language supports efficient communication. Previous efficiency-based approaches have focused on syntactic and semantic structures, but our work shows that the same general approach can capture the tradeoff between communicative pressures in different strategies for lexicalization. Our work therefore suggests that the view that language is shaped to support efficient communication has the potential to explain a wide spectrum of lexicalization strategies in the evolution of the lexicon.

\section{Materials and Methods}
\subsection{Treatment of Data}
We focused on five idealized historical intervals in the past century, setting $t_1 = 1900, 1920,$ ...$, 1980$ and $t_2 = t_1 + 19$. For each interval, we instantiated the components $\mathcal{L}$ and $\encoding$ using form-sense pairs in language-specific WordNets~\cite{fellbaum1998wordnet,bond2013linking}, and we set $\mathcal{C}^*$ to be the concepts encoded in $\encoding$. For tractability, we set the universe $\mathcal{C}$ to be the union of $\mathcal{C}^*$ and concepts encoded in $\mathcal{L}$.
% $\mathcal{C} = \mathcal{C}^* \cup \{c: (c, w) \in \mathcal{L} \text{ for some } w\}$.

We first instantiated $\mathcal{L}$ and $\encoding$ using the English WordNet~\cite{fellbaum1998wordnet} for each interval. Before setting up the components, we standardized word forms in the dataset to facilitate estimation of frequency and word length. We assigned form-sense pairs to $\mathcal{L}$ if their frequencies during $[t_1 - 20, t_2]$ exceeded certain thresholds, based on token frequencies from the Google Ngrams corpus~\cite<English 2020 version;>{michel2011quantitative} and sense frequencies estimated using the Corpus of Historical American English~\cite<COHA;>{davies2002corpus} and a state-of-the-art word sense disambiguation algorithm, EWISER~\cite{bevilacqua-navigli-2020-breaking}. We obtained $\encoding$ by first collecting reuse items and compounds with first citations in $[t_1, t_2]$ according to the Historical Thesaurus of English~\cite{HTE4.2}, and we then took a coarse-grained approach that assumed these items emerged at $t_1$ and have entered the lexicon by $t_2$. Lastly, we processed both $\mathcal{L}$ and $\encoding$ to ensure they are disjoint. We provide a full description of this data processing pipeline in {\it SI Appendix, Section S2.A}.

We instantiated the same components for each interval using French and Finnish WordNets~\cite{sagot2008building,linden2010finnwordnet}. Here, we assumed concepts encoded in the English lexicon $\mathcal{L}$ are also encoded in the French or Finnish lexicon $\mathcal{L}$ for the same interval, and we implemented $\mathcal{L}$ by labeling these concepts with forms that are attested in historical sections of the Google Ngrams corpus~\cite<French 2020 version;>{michel2011quantitative} and the Newspaper and Periodical Corpus of the National Library of Finland~\cite<FNC;>{FNC1_en}. We also assumed the set of emerging concepts $\mathcal{C}^*$ is the same as the English set for the same interval, and we implemented $\encoding$ by pairing each $c \in \mathcal{C}^*$ with one of its French or Finnish forms if the form is in $\mathcal{L}$ or combines forms in $\mathcal{L}$. We provide a full description of this data processing pipeline in {\it SI Appendix, Section S2.B}.

\subsection{Need and Production Distributions} To implement the need distribution and the production policy, we rewrote their product as $p(c, w | \mathcal{L}') = p(w | \mathcal{L}')p(c | w, \mathcal{L}')$ and separately estimated the first and second terms on the right-hand side in each language. 
% Recall that the support of the joint distribution is $\mathcal{L}'$ by assumption.

In the case of English, we estimated the first term using token frequencies in historical texts that appeared during $[t_1, t_2]$ in the English Google Ngrams corpus~\cite{michel2011quantitative}. We defined the first term as $p(w | \mathcal{L}') \propto f_w$, where $f_w$ is the frequency of the form $w$. We estimated the second term in two steps. If $w$ is a unigram, we reused sense frequencies based on text that appeared during $[t_1, t_2]$ in COHA~\cite{davies2002corpus}, so that given sense $c$ with frequency $f_{c, w}$, we have $p(c |w, \mathcal{L}') \propto f_{c, w}$. Otherwise, we used a uniform distribution over concepts in $\mathcal{L}'$ that are labeled by $w$ since our sense disambiguation method did not apply to open compounds. Lastly, we applied add-one smoothing to form-concept pairs in $\mathcal{L}'$. 
% which is normalized by the total frequency of all labels in $\mathcal{L}'$

In the case of French and Finnish, we used the same method to estimate the first term, except we used historical text in the French Google Ngrams corpus~\cite{michel2011quantitative} and the FNC~\cite{FNC1_en}. We estimated the second term based on how an English speaker would infer the distribution via Bayes rule. Specifically, we first assumed that the speaker's prior $p_e(c)$ is the need probability of $c$ estimated from English text during $[t_1, t_2]$ and their likelihood is uniform over all labels of $c$ in the language-specific lexicon $\mathcal{L}'$. We then defined the second term as the posterior, which is given by $p(c | w, \mathcal{L}') \propto p_e(c)$ if $(c, w) \in \mathcal{L}'$ and zero otherwise. We also applied add-one smoothing to form-concept pairs in $\mathcal{L}'$. 
% $q(w | c)$
% Since we did not have access to French and Finnish corpora that are comparable to COHA, 

\subsection{Prototype Model} Here we specify our variant of prototype-based categorization models in Equation~\ref{eq:sim_func}. We first represented each $c \in \mathcal{C}$ by embedding its definition using a state-of-the-art sentence encoder, Sentence-BERT~\cite{reimers2019sentence}, yielding a semantic vector for each concept $c$. Although WordNet definitions were compiled during the past century~\cite{fellbaum1998wordnet}, the encoder was trained on contemporary natural language data, and we view these vectors and the semantic space as an approximation of listener representation of concepts in our target intervals. Throughout this study, we used cosine distance for $d(\cdot, \cdot)$ following~\citeA{reimers2019sentence}.
% This sentence encoder is based on a language model trained on contemporary text, but recent computational work shows that the embedding space derived from this model can nonetheless reflect some historical differences in word meaning~\cite{giulianelli-etal-2020-analysing,giulianelli-etal-2023-interpretable}.

Since the word $w$ is either an existing word in the listener lexicon $\mathcal{L}$ or a combination of existing words, we defined the prototype $\prototype_{w,\mathcal{L}}$ in two parts. If $w$ is in $\mathcal{L}$, we extended the model in~\citeA{reed1972pattern} and defined the prototype as a weighted average of category exemplars, i.e., the concepts encoded by $w$ in $\mathcal{L}$; alternatively, if $w$ is a combination of $\mathrm{N}$ constituents, we used the additive composition function~\cite{mitchell-lapata-2008-vector} to recursively define a composite prototype that combines the prototypes of its constituents~\cite<e.g.,>{smith1988combining}:
\begin{equation}
\prototype_{w,\mathcal{L}} =  
    \begin{cases} 
        & \sum_{c} p(c | w, \mathcal{L}) c \hspace{0.2cm} \text{ if } w \in \mathcal{L} \\
        & \sum_{i} \prototype_{w_{i},\mathcal{L}} \hspace{0.8cm} \text{ else if } w = w_{1}...w_{\mathrm{N}} \in \mathcal{L}^{\mathrm{N}}
    \end{cases}
    \label{prototype}
\end{equation}
Here the expression $w \in \mathcal{L}$ implies $(c, w) \in \mathcal{L}$ for at least one $c \in \mathcal{C}$, and $p(c | w, \mathcal{L})$ was estimated from the relative frequencies of items in $\mathcal{L}$. In {\it SI Appendix, Section S3}, we validated our embedding space and construction of prototypes using datasets of English word similarity and compound meaning predictability.
% human ratings of word meaning change and 
% ashby1995categorization
%  As in our measurement of word length, we simplified our analyses by ignoring the linking constituent in prepositional compounds. 
% talk about sensitivity parameter
% \subsection{Sensitivity Parameter} Here we specify our implementation of the sensitivity parameter $\gamma$ in Equation~\ref{eq:sim_func}. 

Intuitively, the parameter $\gamma$ simulates how much the listener prefers to infer the most transparent interpretation of a word. In {\it SI Appendix, Section S3.C}, we show that the average information loss incurred over communicating existing concepts (i.e., part of the omitted term in Equation~\ref{eq:objective-rewrite}) is minimized when $\gamma \in [15, 20]$. This suggests a reasonable range for $\gamma$ is $(0, 15]$ since if $\gamma$ is too low then the listener does not distinguish among concepts, and if $\gamma$ is too high then the listener will incur high information loss whenever the word form expresses an extended sense. For this reason, in the main text we set $\gamma = 10$. We note that our argument is based on the information loss of existing words, and we leave more fine-grained modeling of the listener distribution for compounds for future work.

\subsection{Estimating the Pareto Frontier} For each tradeoff parameter $\beta = 0, 0.01, ..., 10$, we computed the optimal encoding $\encoding_\beta$ that encodes concepts in $\mathcal{C}^*$ and minimizes Equation~\ref{eq:objective-rewrite}. {We assume that need probabilities are constant with respect to how concepts are encoded. In this case, each novel concept independently contributes to the overall cost in Equation~\ref{eq:objective-rewrite}, and this optimization is equivalent to finding a form $w$ for each $c \in \mathcal{C}^*$ such that $w$ jointly minimizes word length and surprisal for a certain $\beta$.} That is, we want to optimize the following item-level objective over existing words and possible combinations:
\begin{equation}
    L_\beta[w | c, \mathcal{L}] = h(\hat{m}_{w,\mathcal{L}}(c)) + \beta l(w)
    \label{eq:item-level}
\end{equation}
For tractability, we greedily selected a first string $u \in \mathcal{L}$ that optimizes Equation~\ref{eq:item-level}, and we greedily selected a second string $u' \in \mathcal{L}$ or the empty string such that the concatenation of the selected strings minimizes Equation~\ref{eq:item-level}. The final concatenation is the approximately optimal form for $c$ and this form-concept pair is added to $\encoding_\beta$.
% Since $p(c, w | \mathcal{L}')$ was defined as a constant estimated from frequency, each form-concept pair in Equation~\ref{eq:objective-rewrite} independently contributes to the overall cost, and this optimization is equivalent to finding a form $w$ for each $c \in \mathcal{C}^*$ such that $w$ jointly minimizes word length and surprisal for a certain $\beta$

\subsection{Baseline Encodings} We constructed near-synonym and random encodings as baselines for every attested encoding $\encoding$. To construct a near-synonym encoding, we started by constructing a near-synonym set for every form-sense pair in $\encoding$. Suppose that the form contains modifier $w$ and syntactic head $u$; we assumed all English and Finnish compounds are right-headed and all French compounds are left-headed due to their relative prevalence~\cite{lieber2012ie,hyvarinen2019compounds,van2019compounds}. For the constituent $w$, we selected the top $k = 5$ forms, among all existing forms $x \in \mathcal{L}$ that are closest to $w$ in terms of the cosine distance between $\prototype_{w,\mathcal{L}}$ and $\prototype_{x,\mathcal{L}}$ but are not antonyms of $w$ in WordNet. We repeated this procedure for $u$, except we also made sure the possible word classes of the generated constituents overlap with the possible word classes of $u$. The near-synonym set is defined as $\{xy: x\in S_u, y\in S_w\} \cup S_u$, where $S_w$ and $S_u$ are the forms generated for $w$ and $u$, respectively. We then created a near-synonym encoding by replacing the form in each attested form-sense pair with a random sample from its near-synonym set. We constructed random encodings for each $\encoding$ similarly by replacing every attested label with a form uniformly sampled from forms in $\mathcal{L}$ and their combinations. For every $\encoding$, we created $100,000$ near-synonym and random encodings, respectively.
% The near-synonym set is defined as the Cartesian product of forms generated for $w$ and forms generated for $u$, plus the forms generated for $u$.

\subsection{Efficiency Loss} We compared each attested encoding $\encoding$ against the generated baselines by computing their efficiency loss relative to the Pareto frontier following~\citeA{zaslavsky2018efficient}. Given $\encoding_\beta$ for $\beta = 0, 0.01, ..., 10$, the efficiency loss of the encoding $\encoding$ or its corresponding baselines is defined as follows:
\begin{equation}
    \epsilon = \min_\beta \Big( L_\beta[\encoding | \mathcal{L}] - L_\beta[\encoding_\beta | \mathcal{L}] \Big)
    \label{eq:eff_loss}
\end{equation}
This measures the deviation of the encoding $\encoding$ from optimality, or its deviation from the lowest possible amount of information loss given a specific value of average length.

\subsection{Literal and Non-Literal Items}
We classified a reuse item $(c, w)$ as a literal item if the novel concept $c$ is a hyponym of an existing sense of $w$, and otherwise we classified the pair as a non-literal item. We classified a compound item $(c, w)$ as an endocentric compound if $c$ and the head word constitute a literal item, and otherwise we classified the item as an exocentric compound; we made the same assumptions on head positions as in our construction of near-synonyms. Since Princeton WordNet was made to avoid linking a sense and its hyponyms to the same word~\cite{miller1998nouns}, there are few literal reuse items for English ($\text{N} = 3$) and we supplemented attested reuse items for all languages with additional data by replacing the compound in each attested compound-sense pair with its head word. A small number of endocentric compounds with head positions different from our assumption were not used in data augmentation. We provide hypernyms of literal items from Table~\ref{table:data_examples} in {\it SI Appendix, Section S5.B}.

\subsection{Data Availability} All data and code used in analyses are available at \url{https://osf.io/dmgh6}

\section{Acknowledgments}
 
We thank Frank Mollica, Gemma Boleda, Guillaume Thomas, and Michael Hahn for their feedback on the manuscript. We also thank the editor and the reviewers for their constructive feedback. This research is funded partly by U of T-UoM International Research Training Group program. A.X. and Y.X. are supported by Natural Sciences and Engineering Research Council of Canada Discovery Grant RGPIN-2018-05872 and Ontario Early Researcher Award \#ER19-15-050. C.K. is supported by Australian Research Council Grant FT190100200.

\clearpage
\bibliography{references}
\bibliographystyle{apacite}

\end{document}